\documentclass[11pt,a4paper,logo]{lumia}
\usepackage[round,authoryear]{natbib}
\setheadertext{LUMIA Lab}

\usepackage{amsmath,amsfonts,bm}

\def\eqref#1{equation~\ref{#1}}

\def\1{\bm{1}}

\DeclareMathAlphabet{\mathsfit}{\encodingdefault}{\sfdefault}{m}{sl}
\SetMathAlphabet{\mathsfit}{bold}{\encodingdefault}{\sfdefault}{bx}{n}

\usepackage{hyperref}
\usepackage{xurl}
\usepackage{amsmath}
\usepackage{amssymb}
\usepackage{mathtools}
\usepackage{booktabs}
\usepackage{graphicx}
\usepackage{tabularx}
\usepackage{multirow}
\usepackage{xcolor}
\definecolor{bigaired}{RGB}{156,0,0}
\definecolor{color5}{HTML}{006795}
\hypersetup{colorlinks=true,linkcolor=bigaired,citecolor=color5,urlcolor=color5}
\usepackage{algorithm}
\usepackage{algorithmic}
\usepackage{float}
\usepackage{placeins}
\usepackage{needspace}
\usepackage{array}

\title{LoopFold: Parallelizing Recurrent Depth in Looped Transformers}

\author{%
Feichen~Song$^{1}$, Boyi~Zeng$^{1}$, Yiqin~Hao$^{1}$, Shixiang~Song$^{1}$, He~Li$^{1}$,
Yizhou~Yang$^{1}$, Zitong~Wang$^{2}$, Yifan~Liu$^{1}$, Ziwei~He$^{3}$, Zhouhan~Lin$^{1}$\\
$^{1}$Shanghai Jiao Tong University\\
$^{2}$Sun Yat-sen University \quad $^{3}$Shanghai Innovation Institute
}

\newcommand{\method}{LoopFold}
\newcommand{\Backbone}{\mathcal{F}}
\newcommand{\Norm}{\operatorname{RMSNorm}}

\newcommand{\Split}{\operatorname{Split}}
\newcommand{\ConcatB}{\operatorname{Concat}_{\mathrm{batch}}}

\newcommand{\tablecaptiongap}{\vspace{4pt}}

\begin{document}

\begin{abstract}
Looped Transformers increase computation per token by repeatedly applying a
shared backbone, offering a way to improve language modeling and downstream
task performance at a fixed backbone parameter count. However, each loop depends on the preceding loop, so
increasing recurrent depth also increases decoding latency. Existing approaches
include Parallel Loop Transformers, which reduce latency by overlapping loop
computation across token positions, but yield rapidly diminishing marginal
accuracy gains at larger depths. To address this challenge, we introduce \textbf{\method{}},
which folds computation at multiple recurrent depths into parallel batches
for the same token prefix. A lightweight predictor estimates inputs to later loops so
they can start before earlier loops finish. A corrector then refines the
outputs of these later loops using newly available results from earlier
loops. All components are trained jointly using only the final next-token
prediction loss.
In experiments with 1.4B models, deeper \method{} configurations achieve
progressively higher average zero-shot accuracy across eight tasks, rising
from 53.41\% to 56.38\%. Vanilla looped models and PLT show limited gains
over the same depth range. Compared with a nine-loop vanilla model,
\method{} uses five sequential stages and additional parallel computation
to reduce decoding latency by 46\%.

\end{abstract}
\maketitle

\section{Introduction}

Looped Transformers increase computation per token by repeatedly applying a
shared backbone \citep{dehghani2019universal,geiping2025latent}. This
recurrent depth can improve language modeling and downstream task performance
at a fixed backbone parameter count \citep{saunshi2025latent}. However, each loop depends on the preceding loop's
output, so greater depth requires more sequential backbone stages and
increases the wall-clock time needed to generate each token.

Existing approaches reduce decoding latency by overlapping recurrent
computation across token positions. Parallel Loop Transformers (PLT) shift
hidden states by one token between loops, allowing different loop depths to
run concurrently using states from neighboring tokens \citep{wu2025plt}.
Adding loops initially improves accuracy, but the positional mismatch
persists at every loop and the marginal gains quickly diminish at larger
depths \citep{yang2026loopcoderv2}. Recurrent-depth parallel samplers refine
several token drafts in parallel, but reducing the number of recurrent refinement steps
between draft updates trades accuracy for faster decoding
\citep{geiping2025parallel}. The challenge is to start later loops before
their inputs are available while still incorporating the results of
earlier loops.

We introduce \method{} to fold recurrent depth into parallel computation
by batching hidden states from different depths of the same token prefix.
Specifically, a lightweight predictor estimates the inputs to later loops, allowing them to
start before the preceding loops finish. After each batched evaluation, a corrector
uses the newly available earlier-loop results to refine outputs computed from
the estimated inputs. These corrected states feed the next stage, and
the deepest corrected state predicts the next token.

With 1.4B models, \method{} achieves lower PPL and higher mean zero-shot
accuracy as the target depth index increases
(Figures~\ref{fig:compute_latency} and~\ref{fig:depth_accuracy}). At target depth index \(D=9\), \method{} requires only five sequential stages rather than nine, reducing latency by 46\%. It improves accuracy on all eight evaluated tasks, with Avg-8 increasing by 2.20 percentage points relative to the vanilla looped model, at the cost of additional parallel computation.
At matched training backbone-compute budgets of 1.68 and 2.16 ZFLOPs, and
separately at comparable decoding latency, \method{} achieves lower PPL and
higher mean zero-shot accuracy than vanilla looped models. \method{} thus improves the use of recurrent computation in two complementary ways: it can yield better quality at a fixed total backbone budget, and it can exploit available parallelism to perform additional recurrent computation without proportionally increasing decoding latency.

\section{Related Work}

\paragraph{Scaling latent computation.}
Recurrent-depth language models increase computation by repeatedly applying
a shared transition, as in Universal Transformers \citep{dehghani2019universal}.
Such looping can approach the accuracy of deeper models with untied layers
on arithmetic and question-answering tasks \citep{saunshi2025latent}.
PonderLM changes the input passed between loops, feeding a probability-weighted
average of token embeddings into the next loop \citep{zeng2025ponder}.
\citet{geiping2025latent} sample recurrent depths during training to support variable test-time computation. Ouro instead learns an input-dependent exit gate for adaptive recurrent depth \citep{zhu2025ouro}. Hidden Decoding takes a different route,
expanding each token into multiple embedding streams that the backbone
processes together, while retaining their intermediate KV states as context \citep{wechat2026hidden}.

\paragraph{Parallel recurrent computation.}
To shorten the sequential path between loops, PLT and recurrent-depth parallel
samplers overlap computation across token positions. PLT pipelines loop depths
using states shifted across token positions
\citep{wu2025plt}. Recurrent-depth parallel samplers accelerate decoding from pretrained
checkpoints by refining several token drafts concurrently, relying on
per-loop input injection and decodable intermediate states \citep{geiping2025parallel}.
\method{} uses predicted inputs and output correction to run several
backbone loops in parallel for the same token prefix.

\paragraph{Parallel correction methods.}
Classical Parareal uses a fine propagator for accurate parallel local solves
and a cheaper coarse propagator for sequential correction, iterating toward
the fine propagator's sequential solution
\citep{lions2001parareal}. Neural variants replace or parameterize the coarse component with learned networks \citep{lee2021pararealnn}. Related methods parallelize network evaluation
through fixed-point or Newton iterations \citep{song2021parallel,lim2024deer},
with SNLP using structured Jacobian approximations for Transformer layers
\citep{han2026snlp}. \method{} adapts parallel evaluation and correction to a
shared recurrent backbone. Its predictor initializes later-loop inputs once,
allowing several loops to run together, while its corrector incorporates
newly available earlier-loop states into the parallel outputs. Unlike
classical Parareal, which iteratively seeks to recover the sequential fine
solution, \method{} learns a fixed-stage parallel computation optimized
directly for next-token prediction. Its objective is therefore not to match
the sequential recurrent trajectory, but to produce predictive states under
a constrained number of sequential stages.

\begin{figure}[!htb]
\centering
\includegraphics[width=0.90\textwidth,trim=0 6bp 0 6bp,clip]{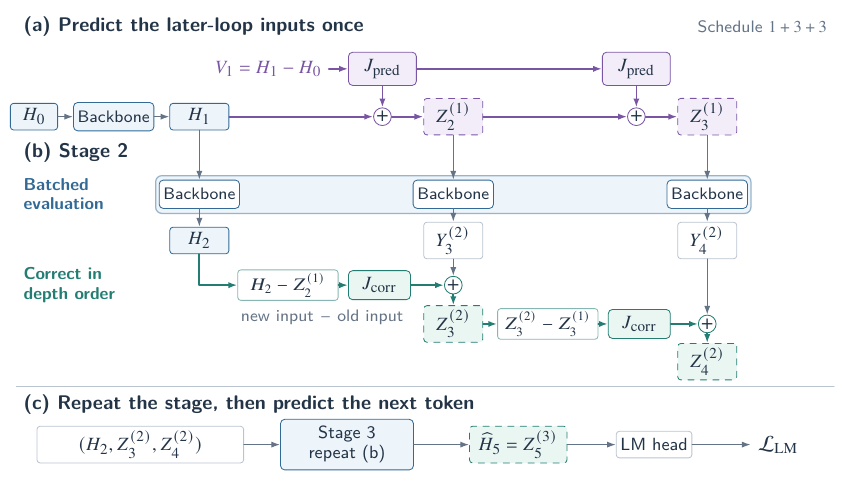}
\caption{\textbf{\method{} with schedule \(1+3+3\).}
Three stages and width 3 reach target depth index \(D=5\) using seven
backbone evaluations. (a) Predict deeper inputs from \(H_1-H_0\).
(b) Evaluate three states together and correct using earlier-loop updates.
(c) Repeat the stage and predict the next token.}

\label{fig:overview}
\end{figure}

\section{Method}
\label{sec:method}

\method{} supplies predicted inputs to later loops so that several recurrent
depths can be evaluated together in a batch. For example,
\(1+3+3\) first computes the state at depth index 1 and predicts provisional
states at indices 2 and 3. The next two sequential stages produce corrected
states at indices \(2,3,4\) and then \(3,4,5\), so the final state has
target depth index \(D=5\) (Figure~\ref{fig:overview}).

Let \(H_0=\operatorname{Embed}(x_{1:L})\in\mathbb{R}^{d\times L}\), with
each column representing one token, and let \(\Backbone\) be the shared
causal Transformer backbone. Its sequential states satisfy
\(H_j=\Backbone(H_{j-1})\). For parallel width \(n\) and total stage count
\(S\), a schedule \(1+n+\cdots+n\) uses one initial backbone evaluation
followed by \(S-1\) stages of \(n\) parallel evaluations.
Each schedule entry counts backbone evaluations in one sequential stage.
A schedule is specified by \((n,S)\); \(D=n+S-1\) is the target depth index
of its final corrected state. Different schedules can share the same \(D\).
Let \(Z_j^{(s)}\) denote the provisional state at depth index \(j\) after stage \(s\).

\subsection{Predicting Inputs for Later Loops}
\label{sec:state_prediction}

After computing \(H_1=\Backbone(H_0)\), we use its state change
\(V_1=H_1-H_0\) as an initial increment along recurrent depth.
A shared token-wise linear map \(J_{\mathrm{pred}}\) repeatedly transforms
this increment, and the accumulated changes initialize deeper states.
At identity initialization, this is simply linear extrapolation of the
first loop's state change:
\begin{equation}
Z_{1+\ell}^{(1)}
=H_1+\sum_{m=1}^{\ell}J_{\mathrm{pred}}^m V_1,
\qquad \ell=1,\ldots,n-1.
\label{eq:stage1}
\end{equation}
Here, \(J_{\mathrm{pred}}^m\) denotes \(m\) successive applications of the
shared predictor map. For \(J_{\mathrm{pred}}=I\), the formula reduces to
\(Z_{1+\ell}^{(1)}=H_1+\ell V_1\). Training adapts the increments to the
language-modeling objective. Prediction runs only once to supply the
\(n-1\) states that join \(H_1\) in the next batch.

\subsection{Parallel Evaluation and Correction}
\label{sec:packed_stages}
\label{sec:output_correction}

At stage \(s=2,\ldots,S\), we batch one sequential state with the provisional
states from the preceding stage and evaluate them together:
\begin{equation}
\begin{aligned}
&\bigl(H_s,Y_{s+1}^{(s)},\ldots,Y_{s+n-1}^{(s)}\bigr) \\
&\qquad =
\Split\!\left[
\Backbone\!\left(
\ConcatB\!\left(
H_{s-1},Z_s^{(s-1)},\ldots,Z_{s+n-2}^{(s-1)}
\right)
\right)
\right].
\end{aligned}
\label{eq:generic_stage_eval}
\end{equation}
Here, \(\ConcatB\) concatenates the \(n\) recurrent states along the batch
dimension, \(\Backbone\) evaluates the \(n\) states in parallel,
and \(\Split\) separates the batched output back into its \(n\) lanes.
All lanes use the same token positions and causal mask, while attention is
computed independently within each lane. The first output advances the
sequential recurrence to \(H_s\); the remaining \(Y_j^{(s)}\) are raw
backbone outputs from the provisional inputs.

The backbone output \(Y_{j+1}^{(s)}\) is stale because it was computed
from the previous estimate of its input. We therefore correct the output
according to how much that input estimate has changed.
A shared token-wise linear map \(J_{\mathrm{corr}}\) translates this input
change into an output adjustment. Starting from \(Z_s^{(s)}=H_s\), we
update the remaining outputs in depth order:
\begin{equation}
Z_{j+1}^{(s)}
=Y_{j+1}^{(s)}+J_{\mathrm{corr}}
\left(Z_j^{(s)}-Z_j^{(s-1)}\right),
\qquad j=s,\ldots,s+n-2.
\label{eq:terminal_solve}
\end{equation}

\begin{table}[!htb]
\caption{\textbf{Backbone computation per token (any model size).}
\(D\): target depth index; \(S\): sequential stages; \(n\): parallel width
after the first loop. Evaluations count all parallel lanes.}
\label{tab:schedule}
\centering
\tablecaptiongap
\small
\setlength{\tabcolsep}{4.5pt}
\renewcommand{\arraystretch}{1.00}
\begin{tabular}{@{}lcccc@{}}
\toprule
Model & \shortstack{Target depth\\index $D$} & \shortstack{Backbone\\evaluations} & \shortstack{Sequential\\stages $S$} & Width $n$ \\
\midrule
Vanilla looped model & $D$ & $D$ & $D$ & 1 \\
\method{} & $n+S-1$ & $1+n(S-1)$ & $S$ & $n$ \\
\bottomrule
\end{tabular}
\end{table}

Because each corrected state is used to correct the next depth,
newly available information propagates sequentially across the parallel outputs.

\subsection{Training Objective and Computational Cost}
\label{sec:schedule_objective}

After \(S\) stages, the deepest state
\(\widehat H_D=Z_{S+n-1}^{(S)}\), with target depth index \(D=n+S-1\), produces
the next-token logits and language-modeling loss:
\begin{equation}
\ell_t=W_{\mathrm{LM}}\Norm(\widehat H_D)_t,
\qquad
\mathcal{L}_{\mathrm{LM}}
=-\sum_{t=1}^{L-1}\log\operatorname{softmax}(\ell_t)_{x_{t+1}}.
\label{eq:lm_objective}
\end{equation}
All components are trained jointly using only this loss, learning
intermediate states for next-token prediction. The predictor and corrector
use separate parameter matrices, each written as \(J=I+\Delta J\), with
\(\Delta J\in\mathbb{R}^{d\times d}\) initialized to zero. Each map is
shared across depths and token positions; the corrector is also shared
across stages. During generation, the prefix stays fixed throughout the
stages, and the next token is selected from the final output.

Counting each parallel lane gives \(1+n(S-1)\) backbone evaluations in
\(S\) sequential stages (Table~\ref{tab:schedule}). The two maps add
\(2d^2\) parameters (8.39M at 1.4B) and \(S(n-1)\) token-wise linear-map
applications per schedule.

\section{Experiments}
\label{sec:experiments}

We evaluate \method{} under three complementary comparison regimes. Matched-training-compute comparisons test whether \method{} can improve model quality at the same total backbone budget. Comparable-latency comparisons test whether additional backbone evaluations can be shifted onto the parallel dimension to improve quality without increasing the number of sequential stages. Finally, matched-target-depth comparisons characterize quality when \method{} reaches the same target depth index with fewer sequential stages. Because these regimes hold different quantities fixed, we report them separately rather than interpreting them interchangeably.

\subsection{Experimental Setup}
\label{sec:experimental_setup}

\noindent\textbf{Models and training.}
We train 1.4B models from scratch on the Pile \citep{gao2020pile} using a
LLaMA-style backbone \citep{touvron2023llama}, a global batch size of 256,
and a common 30B-token budget.
Each schedule is trained separately with the same tokenizer, data order,
and training recipe. The learning rate warms up to \(2\times10^{-4}\)
over the first 1\% of steps, then follows cosine decay to \(2\times10^{-5}\).
The ablations use 410M models trained on 25B tokens.

\noindent\textbf{Baselines.}
A vanilla looped model applies the shared backbone sequentially \(D\) times. Our PLT-\(D\) baselines use cross-loop
pipelining without KV sharing
\citep{wu2025plt}. Hidden Decoding uses 5, 7, or 9 embedding streams per token \citep{wechat2026hidden}.
We omit recurrent-depth parallel samplers because they require per-loop
input injection that our backbone does not use \citep{geiping2025parallel}.

\noindent\textbf{Quality metrics.}
We report Pile validation perplexity (PPL) and zero-shot accuracy on LAMBADA-OpenAI (LAMBADA-OAI) \citep{paperno2016lambada,radford2019language},
ARC-Easy and ARC-Challenge \citep{clark2018arc},
WinoGrande \citep{sakaguchi2020winogrande}, PIQA \citep{bisk2020piqa},
HellaSwag \citep{zellers2019hellaswag}, SciQ \citep{welbl2017sciq}, and
RACE \citep{lai2017race}.

\noindent\textbf{Training computation.}
We report total training backbone computation in ZFLOPs. For the 1.4B models,
one backbone evaluation per original token corresponds to 0.24 ZFLOPs over
the training run. The evaluation count per token is \(1+n(S-1)\) for
\method{}, \(D\) for vanilla looped models and PLT, and the stream count
for Hidden Decoding. This counts computation across all sequential stages,
parallel lanes, and expanded token representations; the learned-map cost
is given in Section~\ref{sec:schedule_objective}.

\noindent\textbf{Decoding latency.}
We measure wall-clock milliseconds per generated token with a 1,024-token
prompt and a 1,024-token greedy continuation. The main comparisons use a
decoding batch size of 1. Section~\ref{sec:batch_scaling} reports aggregate
throughput for batches up to 16. All decoding benchmarks run on the same
GPU in BF16 with tensor parallelism disabled, and timings cover the
complete decoding schedule.

\subsection{Parallel Width at Fixed Stage Count}
\label{sec:schedule_ablation}

\begin{figure}[!htb]
\centering
\includegraphics[width=0.84\linewidth,trim=0 5bp 0 2bp,clip]{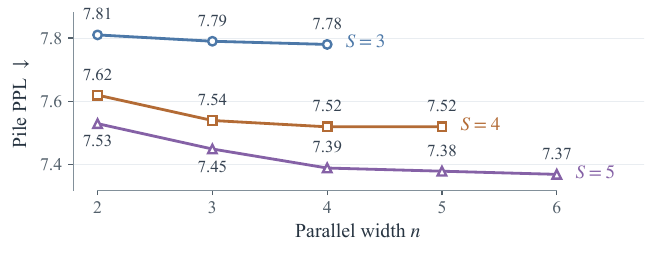}
\caption{\textbf{Parallel width and Pile perplexity in \method{}
(1.4B, 30B training tokens).}
Each curve fixes sequential stages \(S\) and varies width \(n\).
Schedule \(1+n+\cdots+n\) reaches target depth index \(D=n+S-1\).}
\label{fig:schedule_width}
\end{figure}

Parallel width controls how many recurrent states the backbone evaluates
in each batched evaluation. Figure~\ref{fig:schedule_width} varies width while
holding the number of sequential stages fixed at \(S=3,4,5\), testing
whether additional parallel computation improves PPL within the same
sequential-stage budget.

Increasing parallel width at a fixed stage count consistently improves PPL, showing that additional recurrent computation can improve model quality without extending the sequential path. However, the gain per added parallel lane decreases as width grows, indicating progressively lower returns from additional parallel computation.

\begin{figure}[!htb]
\centering
\includegraphics[width=\linewidth,trim=0 6bp 0 0bp,clip]{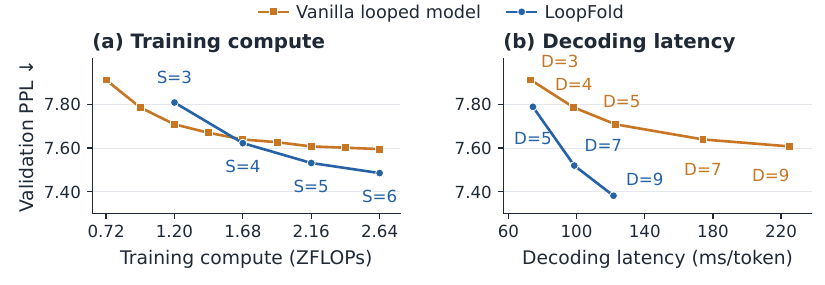}
\caption{\textbf{Pile PPL versus training compute and decoding latency
(1.4B, 30B tokens).}
(a) \method{}: \(n=2\), \(S=3\)--\(6\), ending with \(1+2+2+2+2+2\);
vanilla looped: 3--11 loops.
(b) \method{} uses \(1+3+3\), \(1+4+4+4\), and \(1+5+5+5+5\) at
\(D=5,7,9\). Timing: batch 1, 1,024-token prompts/continuations, greedy
decoding, same GPU, BF16, tensor parallelism disabled.}
\label{fig:compute_latency}
\end{figure}

\subsection{Quality versus Training Compute and Decoding Latency}
\label{sec:main_comparison}
\label{sec:compute_efficiency}
\label{sec:downstream}

At matched training backbone compute, we fix \method{}'s width at 2 and
increase its stage count to compare models across the same training
budgets (Figure~\ref{fig:compute_latency}a and
Table~\ref{tab:downstream_compute}). The vanilla looped model leads at
1.20 ZFLOPs, whereas \method{} achieves lower PPL and higher mean zero-shot
accuracy at 1.68 and 2.16 ZFLOPs. Thus, the parallel schedule can
improve prediction quality within the same training backbone budget while
using fewer sequential stages. \method{} also exceeds PLT and Hidden
Decoding in mean accuracy across all three budgets.

\begin{table}[!tb]
\caption{\textbf{Zero-shot accuracy at matched training backbone compute
(1.4B, 30B Pile tokens).}
Vanilla looped models and PLT use 5/7/9 loops; Hidden Decoding uses 5/7/9
streams per token. \method{} schedules \(1+2+2\), \(1+2+2+2\), and
\(1+2+2+2+2\) correspond to \(D=4,5,6\).
Accuracy is in percent; Avg-8 is the unweighted mean across eight tasks.}
\label{tab:downstream_compute}
\centering
\tablecaptiongap
\small
\setlength{\tabcolsep}{0.7pt}
\renewcommand{\arraystretch}{1.00}
\begin{tabular*}{\linewidth}{@{\extracolsep{\fill}}lcccccccccc@{}}
\toprule
Model & \shortstack{Training\\(ZFLOPs)} & \shortstack{LAMBADA\\OAI} & ARC-E & ARC-C & Wino & PIQA & Hella & SciQ & RACE & Avg-8 \\
\midrule
Vanilla Looped & \multirow{4}{*}{1.20} & \textbf{59.65} & 52.90 & \textbf{28.24} & \textbf{57.14} & \textbf{69.86} & \textbf{50.55} & 78.90 & 32.25 & \textbf{53.69} \\
PLT & & 57.27 & 50.42 & 26.96 & 54.70 & 69.26 & 48.47 & 79.10 & \textbf{33.11} & 52.41 \\
Hidden Decoding & & 54.41 & 48.23 & 26.62 & 52.09 & 68.99 & 43.66 & 75.70 & 30.33 & 50.00 \\
\method{} ($n=2,S=3$) & & 57.60 & \textbf{53.16} & 28.16 & 55.33 & 69.59 & 49.57 & \textbf{79.40} & \textbf{33.11} & 53.24 \\
\midrule
Vanilla Looped & \multirow{4}{*}{1.68} & \textbf{60.14} & 53.32 & \textbf{28.75} & \textbf{57.77} & 70.35 & 51.44 & 77.20 & \textbf{34.64} & 54.20 \\
PLT & & 55.89 & 51.85 & 27.56 & 54.70 & 68.99 & 47.70 & 77.80 & 33.01 & 52.19 \\
Hidden Decoding & & 52.92 & 48.32 & 25.60 & 52.49 & 68.88 & 43.44 & 74.70 & 30.62 & 49.62 \\
\method{} ($n=2,S=4$) & & 58.08 & \textbf{54.80} & \textbf{28.75} & 57.14 & \textbf{71.33} & \textbf{51.80} & \textbf{82.60} & 33.21 & \textbf{54.71} \\
\midrule
Vanilla Looped & \multirow{4}{*}{2.16} & 60.00 & 52.36 & 27.13 & \textbf{57.85} & 70.84 & 51.66 & 79.70 & 33.88 & 54.18 \\
PLT & & 56.84 & 51.05 & 27.22 & 55.41 & 70.29 & 47.72 & 76.30 & 31.96 & 52.10 \\
Hidden Decoding & & 53.19 & 49.20 & 26.54 & 51.78 & 68.23 & 43.65 & 73.40 & 30.62 & 49.57 \\
\method{} ($n=2,S=5$) & & \textbf{60.90} & \textbf{55.39} & \textbf{28.92} & 56.75 & \textbf{71.27} & \textbf{52.79} & \textbf{80.30} & \textbf{34.74} & \textbf{55.13} \\
\bottomrule
\end{tabular*}
\end{table}

At comparable decoding latency, \method{} at target depth indices
\(D=5,7,9\) is compared with vanilla looped models using 3, 4, and 5 loops,
respectively (Figure~\ref{fig:compute_latency}b and
Table~\ref{tab:downstream_latency}). Each pair uses the same number of
sequential stages, allowing \method{} to perform additional recurrent
computation in parallel within a similar time per token. This yields lower
PPL and higher mean accuracy in all three pairs, with gains on all eight
tasks for the two deeper schedules. PLT and Hidden Decoding decode faster
but attain lower mean accuracy.

\begin{table}[!tb]
\caption{\textbf{Zero-shot accuracy and decoding latency (1.4B, 30B Pile tokens).}
\method{} uses \(1+3+3\), \(1+4+4+4\), and \(1+5+5+5+5\) at
\(D=5,7,9\), paired with 3/4/5-loop vanilla looped models.
PLT/Hidden Decoding suffixes count loops/streams per token.
Timing: batch 1, 1,024-token prompts/continuations, greedy decoding, same
GPU, BF16, tensor parallelism disabled. Accuracy is in percent;
Avg-8 is the unweighted mean across eight tasks.}
\label{tab:downstream_latency}
\label{tab:main_downstream}
\centering
\tablecaptiongap
\small
\setlength{\tabcolsep}{0.9pt}
\renewcommand{\arraystretch}{1.00}
\begin{tabular*}{\linewidth}{@{\extracolsep{\fill}}lcccccccccc@{}}
\toprule
Model & \shortstack{Latency\\(ms/token)} & \shortstack{LAMBADA\\OAI} & ARC-E & ARC-C & Wino & PIQA & Hella & SciQ & RACE & Avg-8 \\
\midrule
PLT-5 & 24.93 & 57.27 & 50.42 & 26.96 & 54.70 & 69.26 & 48.47 & 79.10 & 33.11 & 52.41 \\
PLT-7 & 25.35 & 55.89 & 51.85 & 27.56 & 54.70 & 68.99 & 47.70 & 77.80 & 33.01 & 52.19 \\
PLT-9 & 25.16 & 56.84 & 51.05 & 27.22 & 55.41 & 70.29 & 47.72 & 76.30 & 31.96 & 52.10 \\
\midrule
Hidden Decoding-5 & 23.63 & 54.41 & 48.23 & 26.62 & 52.09 & 68.99 & 43.66 & 75.70 & 30.33 & 50.00 \\
Hidden Decoding-7 & 23.95 & 52.92 & 48.32 & 25.60 & 52.49 & 68.88 & 43.44 & 74.70 & 30.62 & 49.62 \\
Hidden Decoding-9 & 23.27 & 53.19 & 49.20 & 26.54 & 51.78 & 68.23 & 43.65 & 73.40 & 30.62 & 49.57 \\
\midrule
Vanilla Looped $D=3$ & 73.11 & 56.36 & 50.21 & 27.65 & \textbf{55.17} & 69.86 & 48.40 & 75.20 & 33.01 & 51.98 \\
\method{} $D=5$ & 74.41 & \textbf{58.01} & \textbf{51.64} & \textbf{27.73} & 54.78 & \textbf{70.24} & \textbf{49.81} & \textbf{81.80} & \textbf{33.30} & \textbf{53.41} \\
\midrule
Vanilla Looped $D=4$ & 98.40 & 58.24 & 52.57 & 28.16 & 54.85 & 70.02 & 50.39 & 77.90 & 34.35 & 53.31 \\
\method{} $D=7$ & 98.66 & \textbf{60.28} & \textbf{54.25} & \textbf{30.03} & \textbf{57.22} & \textbf{70.46} & \textbf{52.67} & \textbf{82.30} & \textbf{34.64} & \textbf{55.23} \\
\midrule
Vanilla Looped $D=5$ & 122.88 & 59.65 & 52.90 & 28.24 & 57.14 & 69.86 & 50.55 & 78.90 & 32.25 & 53.69 \\
\method{} $D=9$ & 121.68 & \textbf{61.61} & \textbf{55.13} & \textbf{29.35} & \textbf{59.75} & \textbf{72.42} & \textbf{54.46} & \textbf{82.30} & \textbf{35.98} & \textbf{56.38} \\
\bottomrule
\end{tabular*}
\end{table}

\subsection{Quality Across Target Depth Indices}
\label{sec:depth_scaling}

\begin{figure}[!htb]
\centering
\includegraphics[width=0.74\textwidth,trim=0 1.5bp 0 4.5bp,clip]{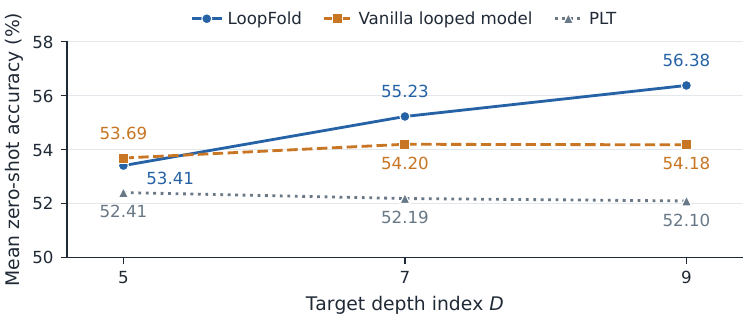}
\caption{\textbf{Mean zero-shot accuracy across target depth indices
(1.4B, 30B Pile tokens).}
At \(D=5,7,9\), \method{} uses \(1+3+3\), \(1+4+4+4\), and
\(1+5+5+5+5\) in 3/4/5 stages; vanilla looped models and PLT use \(D\) loops.
Accuracy is the unweighted mean (\%) across eight tasks.}
\label{fig:depth_accuracy}
\end{figure}

To compare how model quality evolves across deeper configurations,
Figure~\ref{fig:depth_accuracy} reports mean zero-shot accuracy at matched
target depth indices \(D=5,7,9\). The corresponding \method{} schedules
are \(1+3+3\), \(1+4+4+4\), and \(1+5+5+5+5\), which jointly increase
parallel width, sequential stage count, and total backbone computation.

Across these configurations, \method{}'s mean zero-shot accuracy increases
from 53.41\% at \(D=5\) to 56.38\% at \(D=9\), whereas the vanilla looped
and PLT results vary less over the same target depth indices. At \(D=9\), the evaluated \method{} configuration achieves 2.20 percentage points higher Avg-8 accuracy, lower PPL, and 46\% lower decoding
latency than the vanilla looped model, while using additional parallel
backbone computation.

\subsection{Throughput with Larger Decoding Batches}
\label{sec:batch_scaling}

Because \method{} batches recurrent states, decoding more sequences also
enlarges the batch processed by the backbone. Table~\ref{tab:batch_scaling}
reports throughput across decoding batch sizes from 1 to 16. We compare
\method{} \(1+3+3\) (\(D=5\), three sequential stages) with a
five-loop vanilla looped model.

\begin{table}[!htb]
\caption{\textbf{Decoding throughput at \(D=5\) (1.4B, tokens/s).}
\method{} uses \(1+3+3\), with seven backbone evaluations in three stages,
versus five loops in the vanilla looped model. \(B\) is batch size. Timing: 1,024-token
prompts/continuations, greedy decoding, same GPU, BF16, tensor parallelism
disabled. Speed ratio: \method{}/vanilla looped throughput at equal
\(B\); blanks indicate unavailable measurements.}
\label{tab:batch_scaling}
\centering
\tablecaptiongap
\small
\setlength{\tabcolsep}{5pt}
\renewcommand{\arraystretch}{1.00}
\begin{tabular*}{0.96\linewidth}{@{\extracolsep{\fill}}lrrrrr@{}}
\toprule
Model & $B=1$ & $B=2$ & $B=4$ & $B=8$ & $B=16$ \\
\midrule
Vanilla Looped $D=5$ & 8.14 & 16.15 & 31.67 & 65.31 & 130.18 \\
\method{} $D=5$ & 13.44 & 26.24 & 52.26 & 105.35 & 209.12 \\
\midrule
Speed ratio to vanilla looped model & $1.65\times$ & $1.62\times$ & $1.65\times$ & $1.61\times$ & $1.61\times$ \\
\bottomrule
\end{tabular*}
\end{table}

At \(B=1\), \method{} achieves \(1.65\times\) the throughput of the
five-loop vanilla looped model. As batch size increases, \method{}
throughput grows approximately linearly, maintaining stable per-sequence
decoding throughput across the evaluated batch sizes.

\subsection{KV Sharing Across Recurrent Stages}
\label{sec:kv_sharing}

Although \method{} decodes at latency close to a shallower vanilla looped
model, storing an independent KV cache for every recurrent evaluation adds
memory cost. We address this cost in the 1.4B \(1+2+2+2\) model at target
depth index \(D=5\), comparing the default \emph{No KV sharing} policy with
two reuse policies. \emph{First-KV reuse} retains the KV tensors from the
first evaluation at each loop depth and reuses them in later stages,
reducing the seven independent caches to five. To incorporate later-stage
updates for recent tokens, \mbox{\emph{Hard overwrite-64}} additionally
refreshes the most recent 64 cached positions using newly computed
sequential states. Each variant uses the same cache policy in training
and inference.

\begin{table}[!htb]
\caption{\textbf{KV sharing (1.4B, 30B Pile tokens, \(1+2+2+2\), \(D=5\)).}
First-KV reuse retains each depth's first cache; Hard overwrite-64 also refreshes
the latest 64 positions. Storage is relative to one full-backbone cache.
Variants train with their inference cache policies. Timing: batch 1,
1,024-token prompts/continuations, greedy decoding, same GPU, BF16,
tensor parallelism disabled.}
\label{tab:kv_sharing}
\centering
\tablecaptiongap
\small
\setlength{\tabcolsep}{5pt}
\renewcommand{\arraystretch}{1.00}
\begin{tabular*}{0.96\linewidth}{@{\extracolsep{\fill}}llrrr@{}}
\toprule
Variant & Later-stage KV policy & KV storage & \shortstack{Latency\\(ms/token)} & Pile PPL $\downarrow$ \\
\midrule
No KV sharing & independent cache & $7\times$ & \textbf{99.89} & \textbf{7.62} \\
First-KV reuse & reuse all positions & $5\times$ & 103.21 & 7.68 \\
Hard overwrite-64 & overwrite most recent 64 & $5\times$ & 104.94 & 7.63 \\
\bottomrule
\end{tabular*}
\end{table}

First-KV reuse reduces KV storage from \(7\times\) to \(5\times\), with
slightly higher PPL and decoding latency. Hard overwrite-64 retains this
memory saving and recovers most of the PPL loss by refreshing recent
cached states. This refresh adds a small further latency overhead.
\par

\subsection{Contributions of Prediction and Correction}
\label{sec:mechanism_ablations}

To identify how prediction and correction contribute to the quality gains,
we use 410M models with the \(1+4+4+4\) schedule. Component removal tests the
value of each operation, identity maps test the benefit of learning its
transformation, and state measurements examine how provisional states
change across stages.

\subsubsection{Component Removal}

\emph{No prediction} initializes every lane with
the same \(H_1\) while retaining correction. \emph{No correction} keeps
the learned initialization but passes raw backbone outputs directly to
the next stage. All three variants retain the thirteen scheduled backbone
evaluations and are trained separately.

\begin{table}[!htbp]
\caption{\textbf{Component ablations (410M, 25B Pile tokens, \(1+4+4+4\), \(D=7\)).}
\(J_{\mathrm{pred}}\) and \(J_{\mathrm{corr}}\) denote prediction and correction.
No prediction repeats \(H_1\); No correction uses raw backbone outputs.
Variants train separately; \(\Delta\)PPL is the increase over the complete model.}
\label{tab:mechanism_ablations}
\centering
\tablecaptiongap
\small
\setlength{\tabcolsep}{5pt}
\renewcommand{\arraystretch}{1.00}
\begin{tabular*}{0.92\linewidth}{@{\extracolsep{\fill}}lccrr@{}}
\toprule
Variant & Initial states & Output correction & Pile PPL $\downarrow$ & $\Delta$PPL \\
\midrule
\method{} & $J_{\mathrm{pred}}$ & $J_{\mathrm{corr}}$ & \textbf{9.13} & -- \\
No prediction & repeated $H_1$ & $J_{\mathrm{corr}}$ & 9.29 & +0.16 \\
No correction & $J_{\mathrm{pred}}$ & none & 9.47 & +0.34 \\
\bottomrule
\end{tabular*}
\end{table}

Both components improve PPL under the same backbone budget. Removing
correction has the larger penalty, highlighting the value of incorporating
new earlier-loop results into the parallel states.

\subsubsection{Learning the Prediction and Correction Maps}
\label{sec:identity_ablations}

To test the value of learning each map, we hold one map fixed at identity
throughout training while learning the backbone and the other map.
\emph{Identity prediction} therefore uses linear extrapolation, whereas
\emph{Identity correction} adds the input discrepancy directly to the output.

\begin{table}[!htbp]
\caption{\textbf{Learned versus identity maps (410M, 25B Pile tokens,
\(1+4+4+4\), \(D=7\)).}
Each identity variant fixes prediction \(J_{\mathrm{pred}}\) or correction
\(J_{\mathrm{corr}}\) at \(I\) throughout training, while learning the
backbone and the other map.}
\label{tab:identity_ablations}
\centering
\tablecaptiongap
\small
\setlength{\tabcolsep}{5pt}
\renewcommand{\arraystretch}{1.00}
\begin{tabular*}{0.84\linewidth}{@{\extracolsep{\fill}}lccr@{}}
\toprule
Variant & $J_{\mathrm{pred}}$ & $J_{\mathrm{corr}}$ & Pile PPL $\downarrow$ \\
\midrule
\method{} & Learned & Learned & \textbf{9.13} \\
Identity prediction & $I$ & Learned & 9.21 \\
Identity correction & Learned & $I$ & 9.20 \\
\bottomrule
\end{tabular*}
\end{table}

Both identity variants outperform removal of the corresponding component,
showing that linear extrapolation and direct discrepancy addition already
provide useful operations. Learning both maps further lowers PPL, so
adapting these operations to the jointly trained backbone adds value
beyond their identity forms.
\par

\subsubsection{Evolution of Learned States}
\label{sec:correction_convergence}

To examine how the parallel states evolve, Table~\ref{tab:state_closeness}
revisits the three variants in Table~\ref{tab:mechanism_ablations}.
We compare their provisional states with \(H_2,H_3,H_4\) from sequential
rollouts of their own trained backbones on the same input. For each
\(Z_j^{(s)}\), we measure global cosine similarity and relative \(L_2\)
distance:
\begin{equation}
c_j^{(s)}=
\frac{\langle Z_j^{(s)},H_j\rangle_F}
{\lVert Z_j^{(s)}\rVert_F\lVert H_j\rVert_F},
\qquad
r_j^{(s)}=
\frac{\lVert Z_j^{(s)}-H_j\rVert_F}{\lVert H_j\rVert_F}.
\label{eq:state_closeness}
\end{equation}

\begin{table}[H]
\caption{\textbf{Provisional-state evolution (410M, 25B Pile tokens,
\(1+4+4+4\), \(D=7\)).}
Reference states come from each variant's own sequential backbone on the
same input. S1 denotes initialization; arrows mark subsequent stages.
Cosine uses full state tensors; relative \(L_2\) divides the Frobenius error
by the reference state's norm.}
\label{tab:state_closeness}
\centering
\tablecaptiongap
\small
\setlength{\tabcolsep}{2.6pt}
\renewcommand{\arraystretch}{1.00}
\begin{tabular*}{\linewidth}{@{\extracolsep{\fill}}llccc@{}}
\toprule
Variant & Metric & $H_2$ (S1) & $H_3$ (S1$\to$S2) & $H_4$ (S1$\to$S2$\to$S3) \\
\midrule
\multirow{2}{*}{\method{}}
& Cosine $\uparrow$ & 0.922 & 0.774$\to$0.811 & 0.636$\to$0.653$\to$0.823 \\
& Relative $L_2$ $\downarrow$ & 0.394 & 0.642$\to$0.588 & 0.809$\to$0.792$\to$0.573 \\
\midrule
\multirow{2}{*}{No prediction}
& Cosine $\uparrow$ & 0.971 & 0.874$\to$0.924 & 0.643$\to$0.701$\to$0.805 \\
& Relative $L_2$ $\downarrow$ & 0.239 & 0.494$\to$0.389 & 0.777$\to$0.726$\to$0.594 \\
\midrule
\multirow{2}{*}{No correction}
& Cosine $\uparrow$ & 0.506 & 0.517$\to$0.870 & 0.234$\to$0.224$\to$0.414 \\
& Relative $L_2$ $\downarrow$ & 0.866 & 0.857$\to$0.516 & 0.977$\to$1.027$\to$0.920 \\
\bottomrule
\end{tabular*}
\end{table}

The complete model's state at depth index \(j=4\) moves closer to its sequential reference
across stages, while removing correction leaves weaker alignment.
This is consistent with correction carrying earlier-loop information into
deeper provisional states.

Repeated-\(H_1\) initialization starts closer to its own sequential states
at depth indices 2 and 3, yet the complete model achieves better final PPL.
This may indicate that minimizing instantaneous distance to the sequential trajectory is not necessary for good parallel refinement.

\par

\section{Conclusion}

We introduced \method{} to fold recurrent computation into parallel stages
using a jointly trained backbone, predictor, and corrector. Experiments show that
this design can improve language modeling and zero-shot downstream performance
with fewer sequential backbone stages, while the component analyses highlight
the importance of correction. These results suggest that recurrent depth
and sequential decoding depth need not coincide: additional recurrent
computation can be shifted onto the parallel dimension when provisional
states are subsequently corrected.

\method{} reduces sequential depth at the cost of additional backbone computation and thus higher FLOPs. It also increases KV-cache requirements; although KV sharing mitigates this overhead, our experiments show a modest trade-off in quality and/or latency. Reducing these computational and memory costs while preserving the benefits of parallel recurrent evaluation remains an important direction for future work.

\bibliography{paper_references}
\bibliographystyle{plainnat}

\end{document}